\documentclass[letterpaper]{article}
\usepackage{aaai19}  
\usepackage{times}  
\usepackage{helvet}  
\usepackage{courier}  
\usepackage{url}  
\usepackage{latexsym}

\usepackage[utf8]{inputenc}
\usepackage{times}
\usepackage{hyperref}
\usepackage{listings}
\usepackage{footnote}
\usepackage{adjustbox}
\usepackage{xcolor}
\usepackage{wrapfig}
\usepackage{algorithm}
\usepackage{algorithmic}
\usepackage{graphicx,wrapfig,lipsum}
\usepackage{tikz}
\usepackage{amsfonts}
\usetikzlibrary{shapes}
\usepackage{pifont}
\usepackage{microtype}
\usepackage{lscape}
\usepackage{tabulary}
\usepackage{multirow}
\usepackage{caption}
\usepackage{subcaption}
\usepackage{comment}
\usepackage{todonotes}

\usepackage{booktabs}

\usepackage[T1]{fontenc}

\begin{document}

\title{Metrics for Evaluating Quality of Embeddings for Ontological Concepts}

\author{
Faisal Alshargi\\
University of Leipzig \\
Leipzig, Germany \\
alshargi@informatik.uni-leipzig.de \\
\And
Saeedeh Shekarpour\\
University of Dayton \\
Dayton, United States \\
sshekarpour1@udayton.org
\And 
Tommaso Soru\\
University of Leipzig \\
Leipzig, Germany \\
tsoru@informatik.uni-leipzig.de \\
\And 
Amit Sheth\\
Kno.e.sis Center \\
Dayton, United States \\
amit@knoesis.org \\
}

\maketitle

\begin{abstract}
Although there is an emerging trend towards generating embeddings for primarily unstructured data and, recently, for structured data, no systematic suite for measuring the quality of embeddings has been proposed yet.
This deficiency is further sensed with respect to embeddings generated for structured data because there are no concrete evaluation metrics measuring the quality of the encoded structure as well as semantic patterns in the embedding space. 
In this paper, we introduce a framework containing three distinct tasks concerned with the individual aspects of ontological concepts: (i) the categorization aspect, (ii) the hierarchical aspect, and (iii) the relational aspect.
Then, in the scope of each task, a number of intrinsic metrics are proposed for evaluating the quality of the embeddings.
Furthermore, w.r.t. this framework, multiple experimental studies were run to compare the quality of the available embedding models.
Employing this framework in future research can reduce misjudgment and provide greater insight about quality comparisons of embeddings for ontological concepts.
We positioned our sampled data and code at \url{https://github.com/alshargi/Concept2vec} under GNU General Public License v3.0.

\end{abstract}

\section{Introduction}
\label{sec:intro}

Although the Web of Data is growing enormously\footnote{Currently, there are more than 149 billion triples collected from 9,960 data sets of diverse domains, observed on 14 August 2017 at \url{http://stats.lod2.eu/}},
taking advantage of these big interlinked knowledge graphs is challenging.
It is necessary to dispose this valuable knowledge for extrinsic tasks such as natural language processing or data mining.
To do that, the knowledge (i.e. schema level and instance level) has to be injected into current NLP and data mining tools;
by a required transformation from discrete representations to numerical representations (called embeddings).
Hence, the current research trend pays substantial attention to exploring ways of either generating or employing high-quality embeddings in various AI applications such as data mining and natural language processing \cite{word2vec1,word2vec2,glove,rdf2vec}.
However, the recent generation of embedding models on linguistic entities demonstrates higher quality in terms of the proper encoding of structure as well as semantic patterns.
For example, Mikolov \cite{word2vec1,word2vec2} indicated that the vector which separates the embeddings of  \texttt{Man} and \texttt{Woman} is very similar to the vector which separates the embeddings of \texttt{King} and \texttt{Queen}; this geometry disposition is consistent with the semantic relationship. 
In other words, embeddings with high quality hold the semantic and linguistic regularities,
thus, arithmetic operations on them result in semantically consistent results.
Nonetheless, there is still no systematic approach for evaluating the quality of embeddings; therefore, the majority of the state-of-the-art evaluations rely on extrinsic tasks.
An extrinsic evaluation measures the contribution of a given embedding model for a downstream task.
That is, embeddings computed by a model are injected as input features to a downstream task (e.g. sentiment analysis, classification, link prediction tasks). 
Then, changes on performance are compared, whereas an intrinsic evaluation directly investigates syntactic or semantic relationships of linguistic entities in embedding space.
An intrinsic task is typically involved in the use of human judges and requires a query inventory.

Ontological concepts play a crucial role in (i) capturing the semantics of a particular domain, (ii) typing entities which bridge a schema level and an instance level, and (iii) determining valid types of sources and destinations for relations in a knowledge graph.
Thus, the embeddings of the concepts are expected to truly reflect characteristics of ontological concepts in the embedding space.
For example, the hierarchical structure of concepts is required to be represented in an embedding space.
With this respect, an existing deficiency is the lack of an evaluation framework for comprehensive and fair judgment on the quality of the embeddings of concepts.
This paper is particularly concerned with evaluating the quality of embeddings for concepts. 
It extends the state of the art by providing several \emph{intrinsic metrics} for evaluating the quality of the embedding of concepts on three aspects: (i) the categorization aspect, (ii) the hierarchical aspect, and (iii) the relational aspect.
Furthermore, we randomly sampled entities from DBpedia and ran a comparison study on the quality of generated embeddings from Wikipedia versus DBpedia using recent embedding models (those which are scalable in the size of DBpedia).

This paper is organized as follows: the next section reviews the state-of-the-art research about evaluating the quality of embeddings followed by the section presenting the preliminaries and problem statement.
Then, next, we shortly represent popular embedding models.
Section ``evaluation scenarios" proposes three evaluation tasks for measuring the quality of embeddings for ontological concepts.
Each task is equipped with several intrinsic metrics which qualitatively and quantitatively assess quality. 
Moreover, each task exhibits an experimental study on various embedding models.
Last, we discuss the general conclusive observations from our experimental study.

\section{Related Work}
\label{sec:relatedwork}
Recent movement in the research community is more weighted towards learning high quality embeddings or employing embeddings in various applications, and the area of evaluating or benchmarking quality of embeddings in a systematic manner is less studied.
However, there are a few papers about studying evaluation methods for the unsupervised learning of embeddings, but they are limited to unstructured corpora \cite{baroni2014don,baroni2010distributional,evaluation2015}.
Thus, there is a tangible research gap regarding evaluation methods for embeddings learned from a knowledge graph.
To the best of our knowledge, this is the first paper which explores and discusses intrinsic metrics for measuring quality from various dimensions over the embeddings learned out of a knowledge graph. 
Baroni's work \cite{baroni2014don}, extending his previous research \cite{baroni2010distributional}, is pioneering state-of-the-art literature which provides a systematic comparison by extensive evaluation on a wide range of lexical semantics tasks and the application of diverse parameter settings. 
The evaluation metrics which it utilizes are the following. 
\emph{Semantic relatedness:} Asking human subjects to measure the semantic relatedness of two given words on a numerical scale. The query inventory contained both taxonomic relations (e.g. cohyponymy relation king/queen) and broader relationships (e.g. syntagmatic relations amily/planning). 
\emph{Synonym detection:} In this task, multiple choices are displayed for a given target word and the most similar word is detected by comparing the cosine similarity of the target word and all the choices. 
\emph{Concept categorization:} In this task, a set of concepts are given, then the task is to group them into a taxonomic order (e.g., helicopters and motorcycles belong to the vehicle class while dogs and elephants belong to the mammal class). 
\emph{Selectional preference:} Provides a list of noun-verb pairs, then it evaluates the relevance of a noun as a subject or as the object of the verb (e.g., for the given pair people/eat, people receives a high relevance score as the subject of eat and a low score as object).
Another relevant work \cite{evaluation2015} published in 2015 extends Baroni's research by employing new metrics: (i) \emph{analogy:} This task aims at finding a term $x$ for a given term $y$ so that $x : y$ best resembles a sample relationship $a : b$ (e.g. king:queen, man:woman), (ii) \emph{coherence:} 
This task expands the relatedness task to a group evaluation. It assesses the mutual relatedness of a groups of words in a small
neighborhood.

\section{Problem and Preliminaries}
\label{sec:problem}
In this section, we present crucial notions utilized throughout the paper and discuss the main challenge of concern in this paper.

\paragraph{\textbf{Preliminaries.}} An unstructured corpus (i.e. textual data) encompasses a set of words. This set of words is denoted by $\mathbb{W}$ and a given word contained in this set is denoted as $w_i \in \mathbb{W}$.
An embedding model $V^t$ on unstructured data generates a continuous vector representation of $m$ dimensions for each word in set $\mathbb{W}$, formally $V^t: \mathbb{W} \rightarrow \mathbb{R}^m$, where 
$m$ is the length of the latent vector space.
Thus, the word $w_i$ in the space $\mathbb{R}^m$ is represented by the vector $V_{w_i}^t=[x_1^i,x_2^i,...,x_m^i]$.

\paragraph{Knowledge Graph.} A knowledge graph\footnote{In this work, we reference an RDF knowledge graph.}, which is a labeled graph-structured model, empowers data by structure as well as semantics. An RDF knowledge graph $K$ is regarded as a set of triples $(s, p, o) \in R \times P \times (R \cup L)$, where the set of resource $R =C \cup E$ is the union of all RDF entities $E$ and concepts $C$ (from schema or ontology).
Furthermore, $P$ is the set of relations starting from a resource and ending at either a resource or a literal value.
$L$ is the set of literals ($L \cap R = \emptyset$).
We introduce the enhanced set of resources denoted by $R^+$, which is a union of $R^+ =R \cup P$.
Thus, in this context, a given resource $r_i$ can refer to an entity $r_i \in E$, a concept $r_i \in C$ or a property $r_i \in P$. 
An embedding model $V^t$ on a knowledge graph generates a continuous vector representation of $m$ dimensions for each resource (i.e., entity, concept, property) of the set $C \cup E \cup P$, formally denoted as $V^t: R^+=C \cup E \cup P  \rightarrow \mathbb{R}^m$, where 
$m$ is the length of the latent vector space.
Thus, the given resource $r_i$ in the space $\mathbb{R}^m$ is represented by the vector $V_{r_i}^t=[x_1^i,x_2^i,...,x_m^i]$.

\begin{figure*}
\centering
\includegraphics[width=\textwidth]{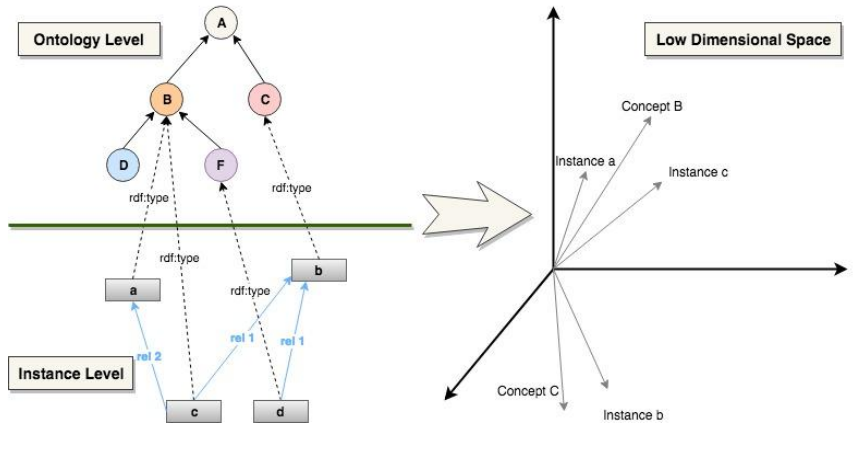}
	\caption{Schematic representation of the vectorization process of a knowledge graph to a low-dimensional space.}
\label{fig:Schematic-Layout}
\end{figure*}

\paragraph{\textbf{Problem Statement.}}
Figure \ref{fig:Schematic-Layout} schematically shows the vectorization process of a knowledge graph to a low dimensional space $V^t:R^+\rightarrow R^m$. A knowledge graph is divided into two levels, (i) an ontology level and (ii) an instance level. 
All the resources from either level (i.e. classes, properties, and entities) are assigned a vector representation in the embedding space. 
The embedding models vary in the \emph{quality} of the generated embeddings. 
The quality of embeddings is attributed to the true reflection of semantics and structural patterns of the knowledge graph in an embedding space.
For example, entities having the same background concept (i.e. common \texttt{rdf:type}) are expected to be clustered close to each other in the embedding space. More importantly, their embedding is expected to be proximate to the embedding of the background concepts (represented in Figure \ref{fig:Schematic-Layout}).
For example, the embeddings of the entities \texttt{dbr:Berlin}, \texttt{dbr:Paris}, \texttt{dbr:London} are expected to be close to the respective concept \texttt{dbo:City} and far from entities such as \texttt{dbr:Barack\_Obama}, \texttt{dbr:Bill\_Clinton} with the respective concept \texttt{dbo:President}.

This paper is particularly concerned with evaluating the quality of embeddings for concepts (i.e. ontological classes) $V^t: C \rightarrow R^m$. 
Generating high quality embeddings for concepts is extremely important since concepts hold the semantics of knowledge graphs. It is expected that these semantics are properly reflected in the embedding space.
For example, the hierarchical semantics (i.e. taxonomic) of concepts is required to be represented in an embedding space.
With this respect, an existing deficiency is the lack of an evaluation framework for comprehensive and fair judgment on the quality of the embeddings of concepts.
While there has recently been a trend for either generating embeddings or employing existing embeddings in various applications, there is not yet a clear framework for intrinsically measuring the quality of embeddings.
This paper contributes in providing several metrics for evaluating the quality of the embedding of concepts from three perspectives: (i) how the embedding of concepts behaves for categorizing their instantiated entities; (ii) how the embedding of concepts behaves with respect to hierarchical semantics described in the underlying ontology; and (iii) how the embedding of concepts behaves with respect to relations.

\section{State-of-the-art Embedding Models}
\label{sec:emModel}
Matrix factorization methods \cite{levy2014neural,glove} and neural networks \cite{word2vec2,word2vec1} are two common approaches for learning dense embeddings for words.
Using neural networks is a recently popularized approach. 
A neural network model starts the learning process with a random embedding for each word, then it iteratively enhances the quality of the embeddings with the criteria that words sharing a common context are more similar and vice versa. 
Thus, adjacent words acquire similar embeddings.
This approach was popularized after the introduction of \emph{word2vec} methods by Mikolov \cite{word2vec2,word2vec1}, where it was shown that the semantic patterns and regularities are well captured by the generated embeddings.
The word2vec methods feature two models for generating embeddings: (i) a skip-gram model and (ii) a continuous bag of words (CBOW) model.
Shortly after, an outperformed model called GloVe \cite{glove} was introduced.
However, all of these models learn embeddings out of the unstructured data.
RDF2Vec \cite{rdf2vec} is a recent state-of-the-art embedding model which learns embeddings out of the knowledge graph.
In the following, we briefly describe each model.

\paragraph{Skip-Gram Model.}
The skip-gram model \cite{word2vec2,word2vec1} learns two separate embeddings for each target word $w_i$, (i) the word embedding and (ii) the context embedding. 
These embeddings are used to compute the probability of the word $w_k$ (i.e. context word) appearing in the neighborhood of word $w_i$ (i.e. target word), $P(w_k|w_i)$.
The skip-gram algorithm (with negative sampling) starts traversing the corpus for any given target word $w_i$. For any occurrence of the target word, it collects the neighboring words as positive samples and chooses $n$ noise samples as negative sampling (i.e., non-neighbor words). 
Eventually, the objective of the shallow neural network of the skip-gram model is to learn a word embedding maximizing its dot product with context words and minimizing its dot products with non-context words.

\paragraph{Continuous Bag of Words (CBOW) Model.}
The CBOW model is roughly similar to the skip-gram model as it is also a predictive model and learns two embeddings for each word (a word embedding and a context embedding).
The difference is that CBOW predicts the target word $w_i$ from the context words as $P(w_i|w_k,w_j)$. 
Thus, the input of the neural network is composed by the context words (e.g. $[w_{i-1},w_{i+1}]$ for the context with length 1); then, the algorithm learns the probability of $w_i$ appearing in the given context.
Although the difference between these two algorithms is slight, they showed different performance in various tasks. 
State-of-the-art evaluations suggest that these algorithms are individually suited to particular tasks.

\paragraph{GloVe Model.}
The GloVe model \cite{glove} is a global log-bilinear regression model for the unsupervised learning of word embeddings.
It captures global statistics of words in a corpus and benefits the advantages of the other two models: (i) global matrix factorization and (ii) local context window methods.
Differently from the skip-gram model, GloVe utilizes the statistics of the corpus, as it relies on global co-occurrence counts.
The GloVe model outperforms the models above for word similarity, word analogy, and named entity recognition tasks.

\paragraph{RDF2Vec Model.}
RDF2Vec \cite{rdf2vec} is an approach for learning embeddings of entities in RDF graphs. 
It initially converts the RDF graphs into a set of sequences using two strategies: (i) Weisfeiler-Lehman Subtree RDF Graph Kernels, and (ii) graph random walks.
Then, word2vec is employed for learning embeddings over these produced sequences.
This approach is evaluated against multiple machine-learning tasks such as instance classification.
Global RDF vector space embeddings \cite{rdf-glove} applies GloVe model on RDF graph and reports the competitive results. 

\paragraph{Translation-based Models.}
The TransE \cite{bordes2013translating} and TransH \cite{wang2014knowledge} models assume that the embeddings of both the entities and relations of a knowledge graph are represented in the same semantic space, whereas the TransR \cite{lin2015learning} considers two separate embedding spaces for entities and relations.
All three approaches share the same principle, for which new relationships can be discovered by translating on hyperplanes.
In other words, summing the vectors of the subject and the predicate, one can obtain an approximation of the vectors of the objects.
An experimental study shows the superiority of the TransR approach~\cite{lin2015learning}.

\paragraph{Other Knowledge Graph Embedding (KGE) Models.} 
Recently, several other approaches have been proposed to embed knowledge graphs.
HolE (Holographic Embeddings) is related to holographic models of associative memory in that it employs circular correlation to create compositional representations~\cite{nickel2016holographic}.
The idea behind DistMult is to consider entities as low-dimensional vectors learned from a neural network and relations as bilinear and/or linear mapping functions~\cite{yang2014embedding}.
ComplEx is based on latent factorization and, with the use of complex-valued embeddings, it facilitates composition and handles a large variety of binary relations~\cite{trouillon2016complex}.
Neural Logic Programming combines the parameter and structure learning of first-order logical rules in an end-to-end differentiable model~\cite{yang2017differentiable}.
All approaches above have shown to reach state-of-the-art performances on link prediction and triplet classification.

\paragraph{\textbf{Excluding of non-scalable  KGE Approaches.}} 
We selected the knowledge graph embedding approaches for the evaluation of our metrics among RDF2Vec, TransE and three of the methods described in the previous subsection (i.e., HolE, DistMult, and ComplEx).
Differently from RDF2Vec, we could not find DBpedia embeddings pre-trained using any of the other approaches online, thus we conducted a scalability test on them to verify their ability to handle the size of DBpedia.
We extracted three nested subsets from DBpedia with a size of $10^4$, $10^5$ and $10^6$ triples, respectively.
The subsets contained instances along with their full Concise Bounded Description\footnote{See \url{https://www.w3.org/Submission/CBD/} for a definition.}, to avoid having incomplete subgraphs.
We launched the algorithms with their default settings on the three subsets on a 64-core Ubuntu server with 256 GB of RAM.
When a run did not terminate converging after 24 hours, we interrupted it.
Surprisingly, while all approaches managed to finish on the $10^4$ and $10^5$ subsets, only ComplEx and DistMult were able to complete the embedding task on the largest one.
However, utilizing a polynomial interpolation of the runtime values, we predicted that none of the approaches would have successfully completed the task on the full DBpedia English dataset -- which has approximately $10^8$ triples -- in reasonable time.
Hence, we decided to select only the more scalable RDF2Vec approach in our evaluation.



\section{Evaluation Scenarios}
\label{sec:eval}

In this section, we introduce three tasks which individually measure the quality of the concept embeddings from three distinct dimensions: (i) the categorization aspect, (ii) the hierarchical aspect, and (iii) the relational aspect.
Furthermore, each task is equipped with multiple metrics for evaluating a given quality dimension from various angles (i.e.  quantitatively, qualitatively, subjectively, and objectively).
\subsection{Task 1: Evaluating the Categorization Aspect of Concepts in Embeddings}
Ontological concepts $C$ categorize entities by typing them, mainly using \texttt{rdf:type}\footnote{Full URI: http://www.w3.org/1999/02/22-rdf-syntax-ns\#type}. 

In other words, all the entities with a common type share specific characteristics. 
For example, all the entities with the type \texttt{dbo:Country}\footnote{\texttt{dbo:} is the prefix for \url{http://dbpedia.org/ontology/}.} have common characteristics distinguishing them from the entities with the type \texttt{dbo:Person}.

In this task, our research question is: How far is the categorization aspect of concepts captured (i.e., encoded) by an embedding model? In other words, we aim to measure the quality of the embeddings for concepts via observing their behaviour in categorization tasks.
To do that, we introduce two metrics which evaluate the categorization aspect in an intrinsic manner.

\paragraph{Dataset Preparation:} From the DBpedia ontology, we selected 12 concepts, which are positioned in various levels of the hierarchy.
Furthermore, for each concept, we retrieved 10,000 entities typed by it (in case of unavailability, all existing entities were retrieved). 
For each concept class, we retrieved 10,000 instances and their respective labels; in case of unavailability, all existing instances were retrieved. 
Then, the embeddings of these concepts as well as their associated instances were computed from the embedding models: (i) skip-gram, and (ii) CBOW and (iii) GloVe trained on Wikipedia and DBpedia\footnote{Using the RDF2Vec package source code available at \url{http://data.dws.informatik.uni-mannheim.de/rdf2vec/} and Glove-RDF2Vec available at \url{https://github.com/miselico/globalRDFEmbeddingsISWC}}.
We created the Wikipedia text corpus by extracting words from the pages of English Wikipedia\footnote{Available at \url{https://dumps.wikimedia.org/enwiki/}.} version 2017/03/01.
We filtered out punctuation, tags, and hyperlink links (textual part of links was remained), then the corpus was turned to lowercase.
Furthermore, the DBpedia English 2015 dataset\footnote{Available at \url{http://downloads.dbpedia.org/2015-10/core-i18n/en/}.} was used to construct our DBpedia corpus; here, we only filtered out datatype properties. 
As hyperparameters for the word2Vec-based approaches, we adopted a window size of 5 and a vector size of 400 for the Wikipedia embeddings, whereas DBpedia embeddings were learned using a window size of 5 and a vector size of 500.
RDF-GloVe was instead set up with a biased random walk based on PageRank, as \cite{rdf-glove} showed to be the best-performing ranking method, with 20 iterations and a vector size of 200.
We used the GloVe word embeddings\footnote{Available at \url{https://nlp.stanford.edu/projects/glove/}.} pre-trained on 840 billion tokens from a common crawl and a vector size of 300.
The length of walks for the RDF2Vec training was set to 8. 
Since in Wikipedia, a given entity might be represented by several tokens, its embedding is calculated as the \textit{average} of the embeddings of all tokens in one setting and the \textit{sum} of the embeddings of all tokens in another setting. 
For instance, the embedding of \texttt{dbr:George\_Washington} in the \textit{sum} setting was computed as \texttt{v(`george') + v(`washington')}\footnote{The benchmarking datasets are available at: \url{https://github.com/alshargi/Concept2vec}}. 

\begin{figure*}[hptb]
\begin{subfigure}[b]{0.5\textwidth}
\includegraphics[width=\textwidth]{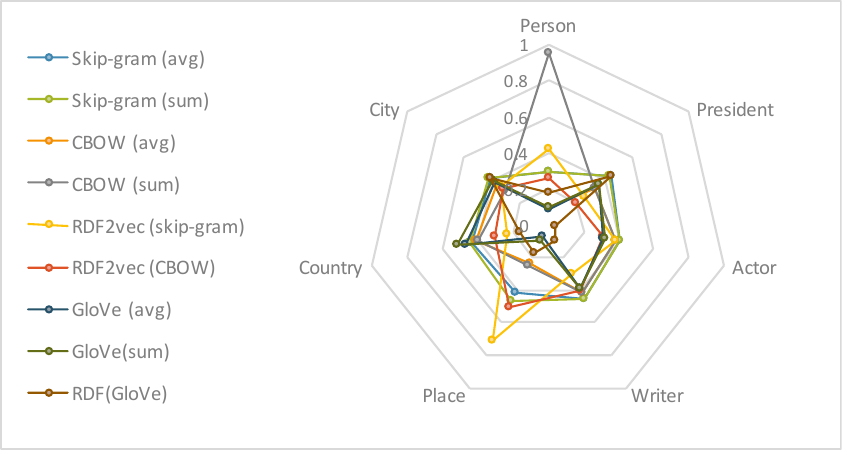}
\caption{\scriptsize Person (Actor,Writer,President), Place (City,Country)}
\label{fig:cat-a}
\end{subfigure}
\begin{subfigure}[b]{0.5\textwidth}
\includegraphics[width=\textwidth]{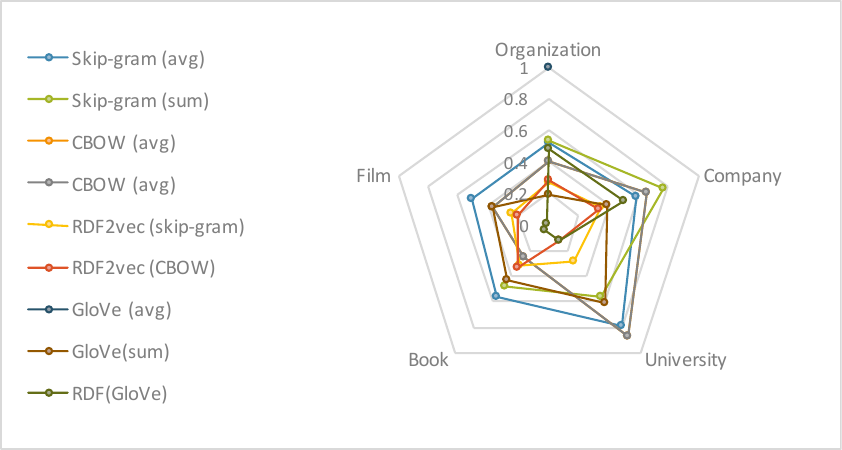}
\caption{\scriptsize Organization(Company,University),Film,Book}
\label{fig:cat-b}
\end{subfigure}

\caption{The categorization score of the twelve concepts for various embedding models trained on Wikipedia and DBpedia.}
\label{fig:Visualization}
\end{figure*}

\paragraph{ \textbf{Categorization metric:}} In the context of unstructured data, this metric aligns a clustering of words into different categories \cite{evaluation2015}. We redefine this metric in the context of structured data as how well the embedding of a concept $c_k$ performs as the background concept of the entities typed by it ($\forall e_i \in c_k$). 
    To quantify this metric, we compute the averaged vector of the embeddings of all the entities having type $c_k$ (represented in Equation \ref{eq:avgVector}) and then compute the cosine similarity of this averaged vector and the embedding of the concept $V_{c_k}$ (formulated in Equation \ref{eq:avgVector2}).
    Please note that throughout the paper $s(V_1,V_2)$ represents the cosine similarity between the two vectors $V_1$ and $V_2$, which is computed as $\frac{V_1.V_2}{|V_1||V_2|}$.

    \begin{equation}
    \label{eq:avgVector}
    \forall e_i \in c_k,
     \overline  V_{c_k}^t=\frac{1}{n}\sum_{i=1}^{i=n} V_{e_i}^t  
    \end{equation}
    
     \begin{equation}
    \label{eq:avgVector2}
      Categorization (V_{c_k}) = s(\overline V_{c_k}^t,V_{c_k}^t)
    \end{equation}
   
\paragraph{Experimental Study:} For each given concept, we measure its categorization score by computing the cosine similarity of its embedding (from a particular model) with the averaged embeddings of its instances. 
Figures \ref{fig:cat-a} and \ref{fig:cat-b} present the results achieved for categorization scores on our underlying data set.
Overall, the skip-gram model outperforms the CBOW model (except in two cases) and GloVe. 
Furthermore, the embeddings learned from Wikipedia outperform the embeddings from DBpedia (again except in two cases).
The other interesting observation of the embedding models is that the categorization score of the concepts positioned in the lower part of the hierarchy (specific concepts) is higher than super concepts (generic concepts). E.g., the categorization score of \texttt{dbo:Place} is lower than its sub-classes \texttt{dbo:City} and \texttt{dbo:Country}.

\paragraph{\textbf{Coherence metric:}} This metric which was introduced in \cite{evaluation2015} measures whether or not a group of words adjacent in the embedding space are mutually related. Commonly, this relatedness task has been evaluated in a \emph{subjective} manner (i.e. using a human judge). 
However, in the context of structured data we define the concept of relatedness as the related entities which share a background concept, a background concept is the concept from which a given entity is typed (i.e. inherited). 
For example, the entities \texttt{dbr:Berlin} and \texttt{dbr:Sana'a} are related because both are typed by the concept \texttt{dbo:City}.
We utilize qualitative as well as quantitative approaches to evaluate the coherence metric. 
In the following, we elaborate on each approach.
\begin{enumerate}
\item \emph{Quantitative evaluation of coherence score:}
Suppose we have a pool of entities with various background concepts and we cluster this pool using the similarity of the embedding of entities. The expectation is that entities with a common background concept are clustered together and, more importantly, the embedding of the background concepts should be the centroid of each cluster.
We follow this scenario in a reverse order. 
For the given concept $c_i$ and the given radius $n$, we find the $n$-top similar entities from the pool (having the highest cosine similarity with $V_{c_i}$). Then, the coherence metric for the given concept $c_i$ with the radius $n$ is computed as the number of entities having the same background concept as the given concept; formally expressed as:

\begin{equation}
  coherence(V_{c_i},n)= \frac{\{\# e_i | e_i \in c_i \}}{n}  
\end{equation}

\item \emph{Qualitative evaluation of coherence score:}
Commonly, the coherence metric has been evaluated by a qualitative approach.
For example, \cite{Turian2015} uses a two-dimensional visualization of word embeddings for measurement by human judges in the relatedness task.
Apart from visualization, another way of qualitative evaluation is providing samples of grouped entities and a concept to a human subject to judge their relatedness.
\end{enumerate}

\begin{figure*}[hptb]
\vspace{-0.5cm}
\begin{subfigure}[b]{0.5\textwidth}
\includegraphics[width=\textwidth]{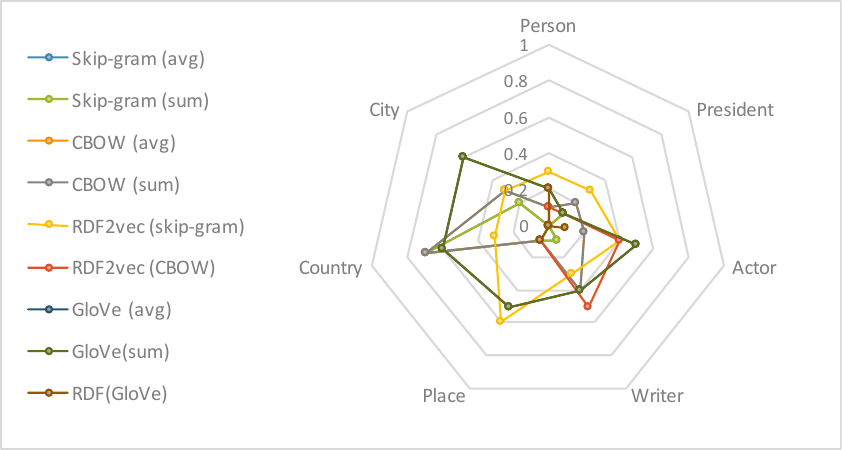}
\caption{\scriptsize Person (Actor,Writer,President), Place (City,Country)}
\label{fig:a}
\end{subfigure}
\begin{subfigure}[b]{0.5\textwidth}
\includegraphics[width=\textwidth]{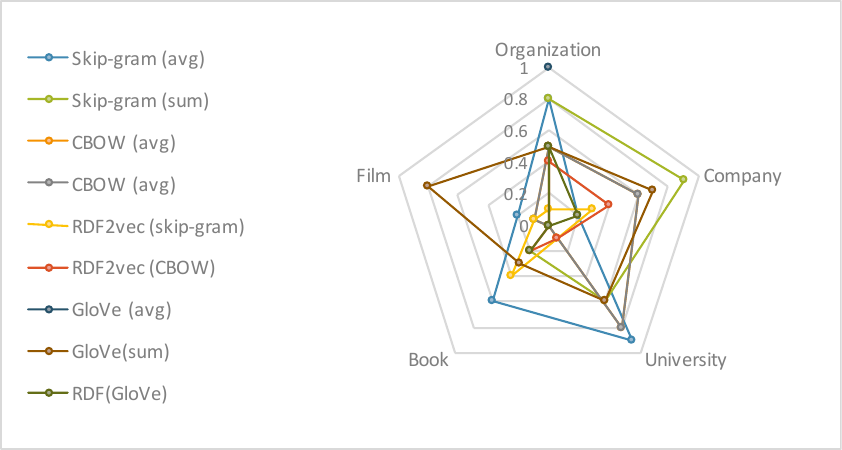}
\caption{\scriptsize Organization(Company,University),Film,Book}
\label{fig:b}
\end{subfigure}
\begin{subfigure}[b]{0.5\textwidth}
\includegraphics[width=\textwidth]{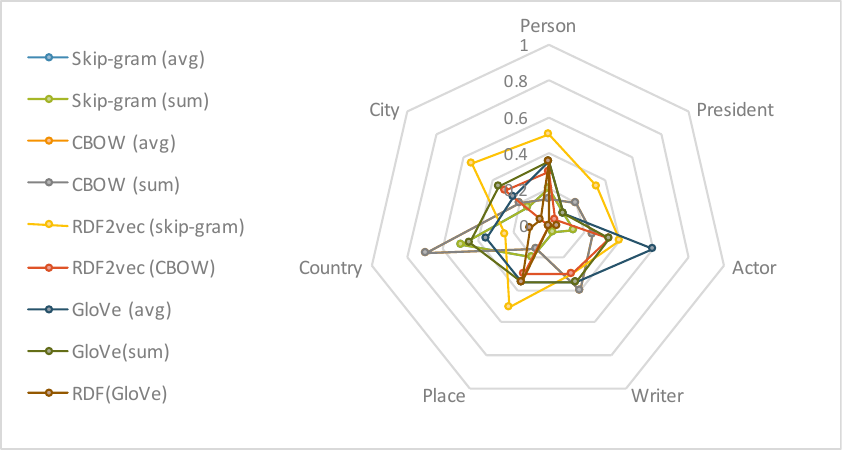}
\caption{\scriptsize Person (Actor,Writer,President), Place (City,Country)}
\label{fig:c}
\end{subfigure}
\begin{subfigure}[b]{0.5\textwidth}
\includegraphics[width=\textwidth]{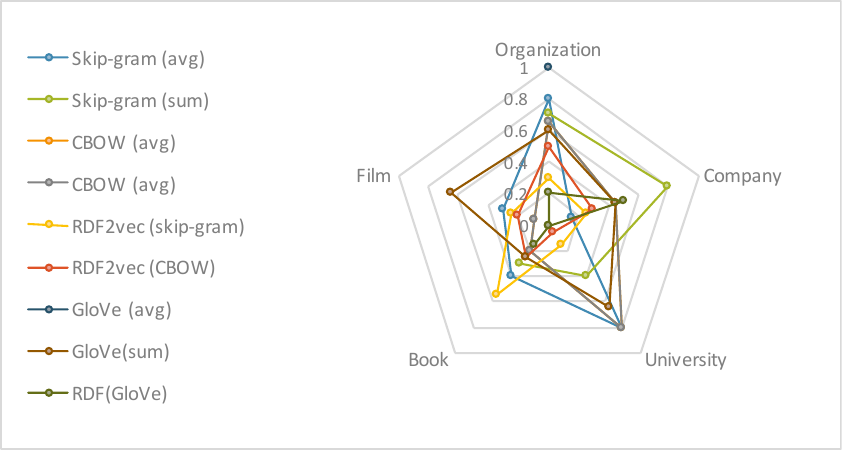}
\caption{\scriptsize Organization(Company,University),Film,Book}
\label{fig:d}
\end{subfigure}

\caption{The coherence score of the twelve concepts with a radius of 10 for (a-b) and a radius of 20 for (c-d).}
\label{fig:Coh-Vi}

\end{figure*}

\paragraph{Experimental Study.} In this experiment, we quantitatively measure the coherence score.
To do that, we initially have to prepare a proper data set.
We reuse the previous dataset with a few modifications.
E.g., for each concept, we sampled a batch containing 20 entities.
Then, all of these batches are mixed up as a single data set. 
This dataset is utilized in the whole of this experiment.
To measure the coherence score for every given concept, we computed the cosine similarity of the given concept and the whole of the entities included in our dataset (which is a mix of entities with various types). 
Then, we list the top-n entities (i.e. $n$ is the radius) which are the closest entities to the given concept (using cosine similarity over the associated embeddings). 
The coherence score is computed by counting the number of entities out of the top-n entities which are typed by the given concept.  
For example, for the given concept \texttt{dbo:Actor}, if three entities out of the top-10 closest entities are not of the type \texttt{dbo:Actor} (e.g. \texttt{dbr:Berlin}), then the coherence score of \texttt{dbo:Actor} is 0.7.
Figure \ref{fig:Coh-Vi} shows the results achieved for the coherence scores for the 12 concepts of our dataset.

The radius value in the experiments showed in Figures \ref{fig:a} and \ref{fig:b} is 10 and in Figures \ref{fig:c} and \ref{fig:d} is 20.
Within the longer radius (i.e. $n=20$), the coherence scores are increased (except for a few cases) especially for the super concepts (e.g. \texttt{Person}, \texttt{Place} and \texttt{Organisation}).
With respect to the models trained on Wikipedia, the GloVe model commonly outperformed
while regarding the models trained on DBpedia on average the skip-gram model performs better.
Generally, the embeddings learned from Wikipedia have the higher coherence scores than the embeddings trained on DBpedia.

\subsection{Task 2: Evaluating Hierarchical Aspect of Concepts in Embeddings}

There is a relatively longstanding research for measuring the similarity of two given concepts $s(c_i,c_j)$ either across ontologies or inside a common ontology \cite{maedche2002measuring,shvaiko2005survey,batet2013semantic}.
Typically, the similarity of concepts is calculated at the lexical level and at the conceptual level.
However, our assumption here is that our underlying knowledge graph has a well-defined ontology as the background semantics.
The concepts of the given ontology are positioned in a hierarchical order and share various levels of semantics.
We present three metrics which can be employed for evaluating the embeddings of concepts with respect to the hierarchical structure and the semantics.

\paragraph{\textbf{Absolute semantic error.}}
We introduce the metric \emph{absolute semantic error} which quantitatively measures the quality of embeddings for concepts against their semantic similarity.
The semantic similarity between the two given concepts $c_i$ and $c_j$ is denoted by $s'(c_i,c_j)$ and can be measured by an state-of-the-art methodology \cite{gan2013ontology,maedche2002measuring,shvaiko2005survey}. 
Ideally, this similarity score should be approximate to the similarity score of embeddings corresponding to those concepts denoted by $s(V_{c_i}^t,V_{c_j}^t)$ (please note that this score is calculated by cosine similarity).
Therefore, this expected correlation can be formally represented as $s'(c_i,c_j) \approx s(V_{c_i}^t,V_{c_j}^t)$.
For example, the semantic similarity between the two concepts $c_1=$ \texttt{dbo:President} and $c_2=$ \texttt{dbo:City} is almost zero; so it is expected that their vectors reflect the similar pattern as $s(V_{c_1}^t,V_{c_2}^t) \approx 0$.
An intuitive methodology for measuring semantic similarity between two concepts is to utilize the distance between them in the hierarchical structure \cite{taieb2014ontology}.
Because, intuitively, the concepts which are placed closer in the hierarchy are more similar.
In contrast, concepts placed further from each other are more dissimilar. Thus, by increasing the length of the path between two concepts in the hierarchy, their dissimilarity is increased.
However, independent of the kind of methodology employed for computing the semantic similarity score, the absolute semantic distance $\Delta$ is computed as the difference between the semantic similarity score $s'$ and the similarity score of embeddings $s$, which is formally represented in Equation \ref{eq:absoluteerror}. The higher the value of $\Delta$, the lower the quality of the embeddings and vice versa. It is formally calculated as :
 \begin{equation}
    \label{eq:avgVector}
    \Delta (c_i,c_j) =  | s'(c_i,c_j) - s(V_{c_i}^t,V_{c_j}^t) |
    \end{equation}
    
\paragraph{\textbf{Semantic Relatedness metric.}} We tune this metric from \cite{baroni2014don,evaluation2015} for knowledge graphs by exchanging words for concepts. 
Typically, this metric represents the relatedness score of two given words.
In the context of a knowledge graph, we give a pair of concepts to human judges (usually domain experts) to rate the relatedness score on a predefined scale, 
then, the correlation of the cosine similarity of the embeddings for concepts is measured with human judge scores using Spearman or
Pearson.

\paragraph{\textbf{Visualization.}} The embeddings of all concepts of the knowledge graph can be represented in a two-dimensional visualization.
This approach is an appropriate means for qualitative evaluation of the hierarchical aspect of concepts. The visualizations are given to a human who judges them to recognize patterns revealing the hierarchical structure and the semantics.

\begin{figure*}[hptb]
\begin{subfigure}[b]{0.5\textwidth}
\includegraphics[width=\textwidth]{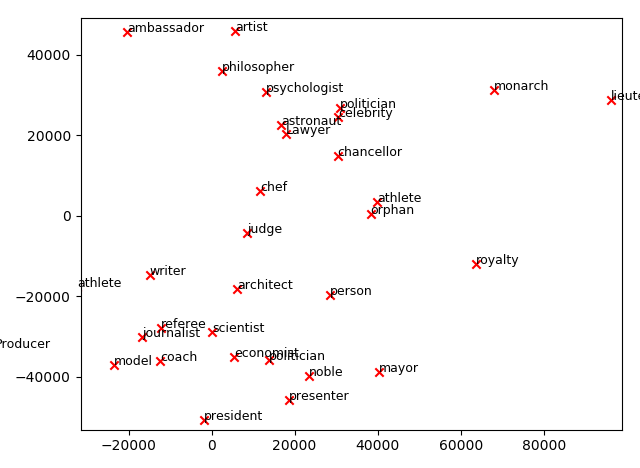}
\caption{\scriptsize DBpedia-CBOW: dbo:Person and its subclasses.}
\label{fig:DB-Person-CBOW}
\end{subfigure}
\begin{subfigure}[b]{0.5\textwidth}
\includegraphics[width=\textwidth]{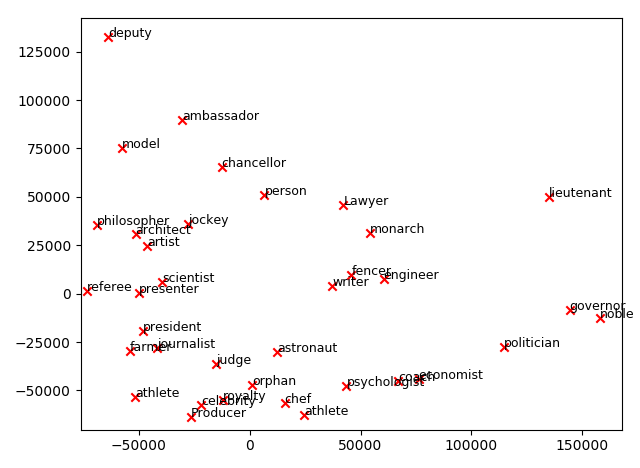}
\caption{\scriptsize DBpedia-Skip-gram: dbo:Person and its subclasses.}
\label{fig:DB-Person-SG}
\end{subfigure}
\begin{subfigure}[b]{0.5\textwidth}
\includegraphics[width=\textwidth]{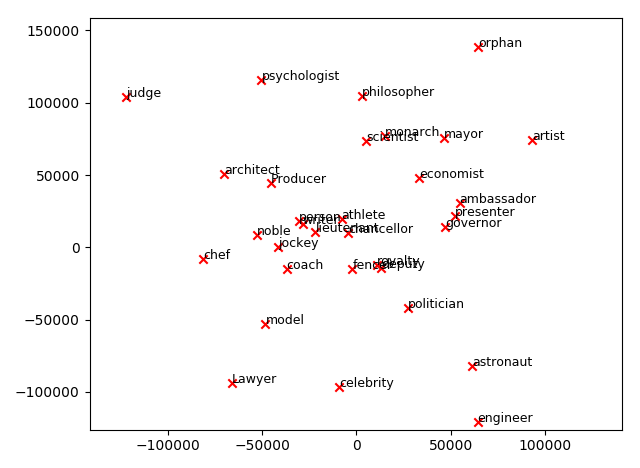}
\caption{\scriptsize Wikipedia-CBOW: dbo:Person and its subclasses.}
\label{fig:wiki-Person-CBOW}
\end{subfigure}
\begin{subfigure}[b]{0.5\textwidth} 
\includegraphics[width=\textwidth]{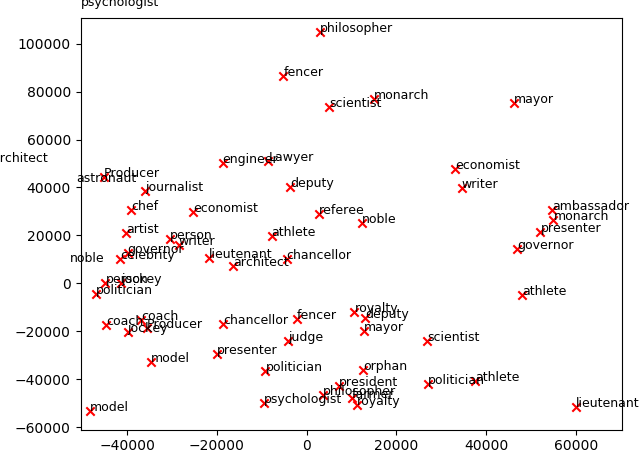}
\caption{\scriptsize WikiPedia-Skip-gram: dbo:Person and its subclasses.}
\label{fig:wiki-Person-SG}
\end{subfigure}
\begin{subfigure}[b]{0.5\textwidth}
\includegraphics[width=\textwidth]{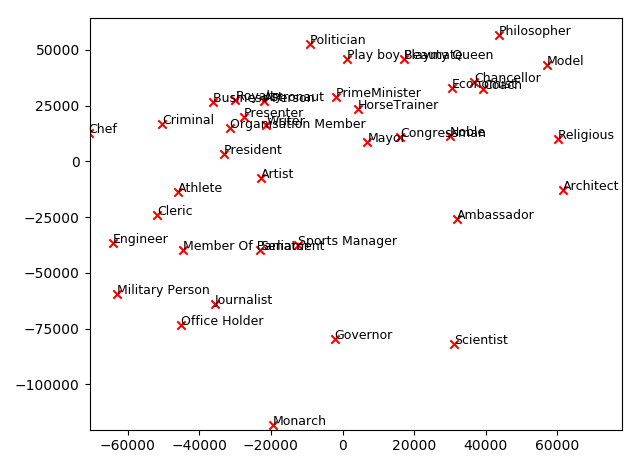}
\caption{\scriptsize Wikipedia-GloVe: dbo:Person and its subclasses.}
\label{fig:wiki-Person-G}
\end{subfigure}
\begin{subfigure}[b]{0.5\textwidth}
\includegraphics[width=\textwidth]{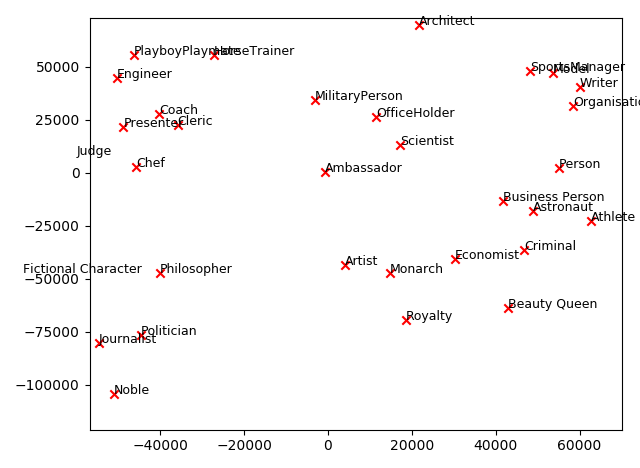}
\caption{\scriptsize DBpedia-GloVe: dbo:Person and its subclasses.}
\label{fig:db-Person-G}
\end{subfigure}

\caption{Two-dimensional visualization of dbo:Person branches of the DBpedia hierarchy.}
\label{fig:Visualization-person}
\end{figure*}


\begin{figure*}[hptb]

\begin{subfigure}[b]{0.5\textwidth}
\includegraphics[width=\textwidth]{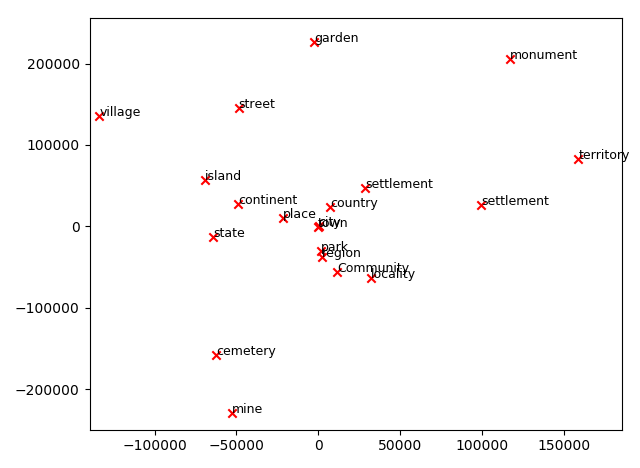}
\caption{\scriptsize DBpedia-CBOW: dbo:Place and its subclasses.}
\label{fig:DB-Place-CBOW}
\end{subfigure}
\begin{subfigure}[b]{0.5\textwidth}
\includegraphics[width=\textwidth]{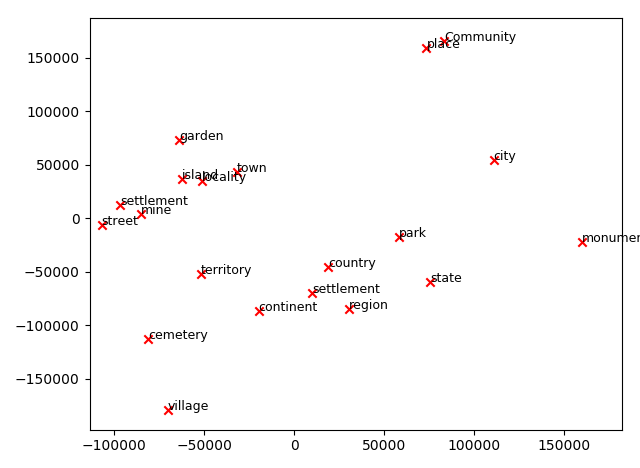}
\caption{\scriptsize DBpedia-Skip-gram: dbo:Place and its subclasses.}
\label{fig:DB-Place-SG}
\end{subfigure}
\begin{subfigure}[b]{0.5\textwidth}
\includegraphics[width=\textwidth]{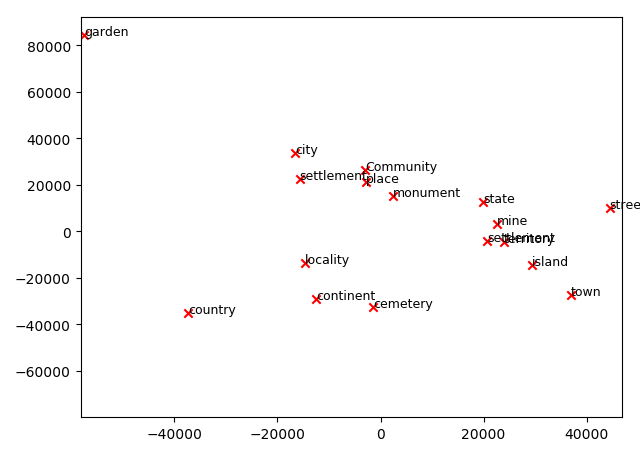}
\caption{\scriptsize Wikipedia-CBOW: dbo:Place and its subclasses.}
\label{fig:Wiki-Place-CBOW}
\end{subfigure}
\begin{subfigure}[b]{0.5\textwidth}
\includegraphics[width=\textwidth]{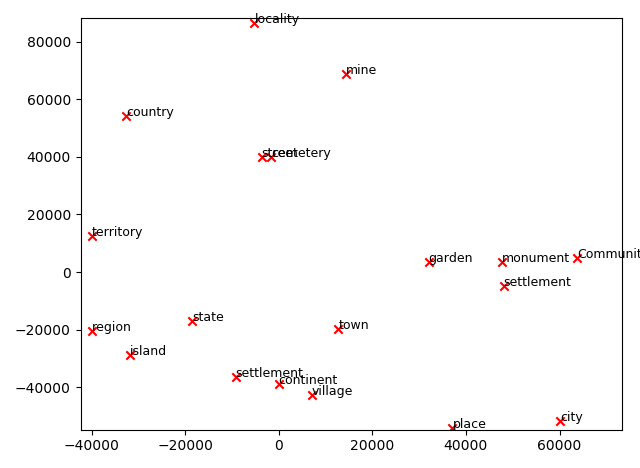}
\caption{\scriptsize Wikipedia-Skip-gram: dbo:Place and its subclasses.}
\label{fig:wiki-Place-SG}
\end{subfigure}
\begin{subfigure}[b]{0.5\textwidth}
\includegraphics[width=\textwidth]{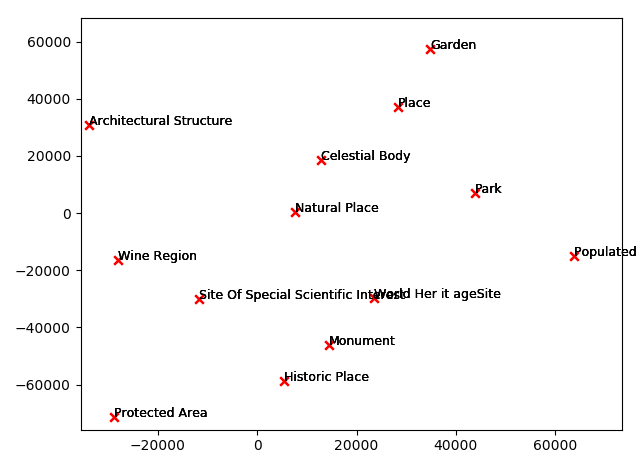}
\caption{\scriptsize Wikipedia-GloVe: dbo:Place and its subclasses.}
\label{fig:wiki-Place-G}
\end{subfigure}
\begin{subfigure}[b]{0.5\textwidth}
\includegraphics[width=\textwidth]{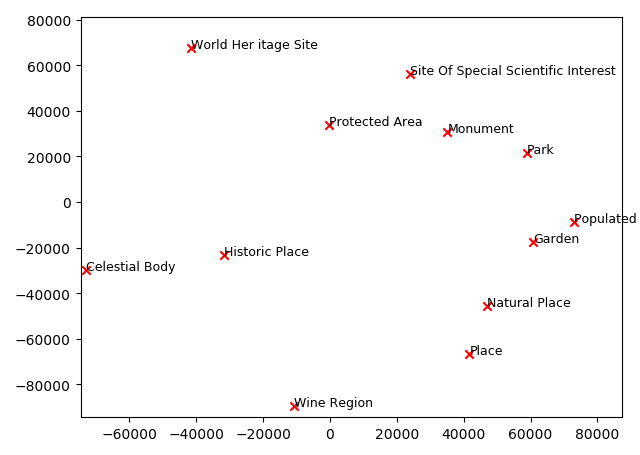}
\caption{\scriptsize DBpedia-GloVe: dbo:Place and its subclasses.}
\label{fig:db-Place-G}
\end{subfigure}

\caption{Two-dimensional visualization of dbo:Place branches of the DBpedia hierarchy.}
\label{fig:Visualization-place}
\end{figure*}

\paragraph{Experimental Study:}
We chose three high level concepts from the DBpedia ontology\footnote{\url{http://mappings.dbpedia.org/server/ontology/classes/}} with their direct children (i.e., linked by \texttt{rdfs:subClassOf}).
In addition, for each of these three concepts, two more concepts placing lower (in the hierarchy) were chosen along with their direct children. 
Herein, for brevity we only name the main concepts chosen. Respectively, the concepts chosen are  (i) \texttt{dbo:Person} with the two sub-concepts \texttt{dbo:Athlete} and \texttt{dbo:Politician},  (ii) \texttt{dbo:Place} with the two sub-concepts \texttt{dbo:Settlement} and \texttt{dbo:PopulatedPlace}. 
To perform the visualization task, we used t-SNE \cite{tSNE} package to reduce the high-dimensional embeddings to two-dimensional embeddings.

Figures \ref{fig:Visualization-person} and \ref{fig:Visualization-place} illustrate the two-dimensional visualizations of the embeddings for the chosen sections of the DBpedia hierarchy\footnote{Please note that the scale of all the diagrams is unified.}.

This visualization facilitates comparison on the quality of the embeddings generated by the GloVe model versus the skip-gram and CBOW models and, furthermore, the effect of the knowledge graph in front of the unstructured data (DBpedia versus Wikipedia).
Figures \ref{fig:DB-Person-CBOW},\ref{fig:DB-Person-SG},\ref{fig:wiki-Person-CBOW}, \ref{fig:wiki-Person-SG}, \ref{fig:wiki-Person-G} and \ref{fig:db-Person-G} represent the 2D visualizations of the embeddings for the concept \texttt{dbo:Person} and its chosen sub-concepts.
Please note that all of these concepts have a taxonomic relationship (i.e. either parental or sibling) with each other.
Generally, the GloVe model on DBpedia and Wikipedia, in comparison to other settings, demonstrates regularities such as (i) having a denser representation between the concepts,
(ii) the centrality of the super-class \texttt{dbo:Person} is higher, (iii) the closeness of the embeddings such as \texttt{dbo:Monarch} and \texttt{dbo:royalty} indicates greater shared semantics compared with other siblings.
Figures \ref{fig:DB-Place-CBOW}, \ref{fig:DB-Place-SG}, \ref{fig:Wiki-Place-CBOW}, \ref{fig:wiki-Place-SG}, \ref{fig:wiki-Place-G} and \ref{fig:db-Place-G} display the 2D visualizations of the embeddings for the concept \texttt{dbo:Place} and its chosen sub-concepts.
The observations which can be concluded are as follows:
(i) the embeddings generated from Wikipedia are denser than the embeddings from DBpedia, (ii) the centrality of the embedding of the concept \texttt{dbo:Place} in GloVe and CBOW models is higher in both Wikipedia and DBpedia, (iii) generally the closeness of the embeddings in CBOW model (either on Wikipedia or DBpedia) is compatible with the siblings sharing higher semantics such as \texttt{dbo:Community-dbo:Locality} or \texttt{dbo:City-dbo:Town} in Figure \ref{fig:DB-Place-CBOW} or \texttt{dbo:Park-dbo:Garden} in Figure \ref{fig:Wiki-Place-CBOW}.

\subsection{Task 3: Evaluating Relational Aspect of Concepts in Embeddings}
There are various applications in information extraction, natural language understanding, and question answering involved in extracting either implicit or explicit relationships between entities \cite{ramakrishnan2006framework,heim2010interactive,augenstein2012lodifier}.
A major part of evaluating the state-of-the-art approaches for relation extraction is the \emph{validation task} as whether or not the inferred relation is compatible with the type of entities engaged.
For example, the relation \texttt{capital} is valid if it is recognized between entities with the types \texttt{country} and \texttt{city}.
This validation process in a knowledge graph is eased by considering the axioms \texttt{rdfs:domain} 
and \texttt{rdfs:range} of the schema properties and \texttt{rdf:type} of entities.
The expectation from embeddings generated for relations is to truly reflect compatibility with the embeddings of the concepts asserted in the domain and range.
With this respect, we present two metrics for evaluating the quality of the embeddings for concepts and relations.

\paragraph{ \textbf{Selectional preference}} This metric presented in \cite{baroni2014don,baroni2010distributional} assesses the relevance of
a given noun as a subject or object of a given verb (e.g. people-eat or city-talk).
We tune this metric for knowledge graphs as pairs of concept-relation which are represented to a human judge for the approval or disapproval of their compatibility.

\paragraph{ \textbf{Semantic transition distance}}
The inspiration for this metric comes from \cite{word2vec1,word2vec2}, where Mikolov demonstrated that capital cities and their corresponding countries follow the same distance. 
We introduce this metric relying on an objective assessment.
This metric considers the relational axioms (i.e. \texttt{rdfs:domain} 
and \texttt{rdfs:range}) in a knowledge graph. Assume that the concept $c_i$ is asserted as the domain of the property $p_i$ and the concept $c_j$ is asserted as its range.
It is expected that the sum of the embeddings of the $c_i$ and $p_i$ conducts to the embeddings of the concept $c_j$.
In other words, the transition distance denoted by $Tr$ measures the similarity (e.g. cosine similarity) of the destination embedding $V_{c_j}$ and the conducted point (via $V_{c_i}+V_{p_j}$), formally expressed as:

\begin{equation}
    \label{eq:avgVector}
   Tr (c_i+p_i,c_j) =  s(V_{c_i}+V_{p_j} , V_{c_j}) 
    \end{equation}

\begin{table*}[hptb]

\scriptsize
\centering
  \begin{tabular}{|l|l|l|l|l|l|l|l|l|}
    \toprule
    \multirow{2}{*}{\textbf{Relation}} &
      \multicolumn{2}{c}{} &
      \multicolumn{3}{c}{\textbf{DBpedia}} &
      \multicolumn{3}{c|}{\textbf{Wikipedia}} \\
    & \textbf{Domain} & \textbf{Range} & skip-gram & CBOW & GloVe & Skip-gram & CBOW & GloVe\\
    \midrule
    
    \texttt{spouse} & \texttt{Person} & \texttt{Person} &0.498  &  0.228 & 0.748 &0.834  & \textbf{0.863} &\textbf{0.863} \\  
    \hline
    \texttt{capital} & \texttt{Populated P.} & \texttt{City} & 0.592    &0.211  &-0.032 &0.532  &0.389 & \textbf{0.676} \\ 
    \hline
   \texttt{starring} & \texttt{Work} & \texttt{Actor} & 0.303 &0.138  &0.231 & 0.563     &0.453 & \textbf{0.656} \\
    \hline
    \texttt{largestCountry} & \texttt{Populated P.} & \texttt{Populated P.} &0.702  & 0.766 & 0.642& \textbf{0.878}  &0.865 &0.863 \\
    \hline
   \texttt{director} & \texttt{Film} & \texttt{Person} &0.15  &0.072  &0.014 &0.173  & 0.056  & \textbf{0.257} \\
    \hline
   \texttt{child} & \texttt{Person} & \texttt{Person} &0.461   &0.173  &  0.71& 0.857  & \textbf{0.869} & 0.866 \\
    \hline
   \texttt{writer} & \texttt{Work} & \texttt{Person} & 0.193 & 0.022 &-0.049 &0.276   & 0.086 &   \textbf{0.46} \\
    \hline
   
   \texttt{school} & \texttt{Person} & \texttt{Institution} & 0.279  &0.262  & 0.087&  0.455 & 0.521 & \textbf{0.541} \\
    \hline
   \texttt{translator} & \texttt{Work} & \texttt{Person} &  0.24  & 0.179  & -0.012& 0.254  &0.095 & \textbf{0.394} \\
    \hline
   \texttt{producer} & \texttt{Work} & \texttt{Agent} & 0.234   &0.006  & 0.212 &  0.229 &0.131 & \textbf{0.357}   \\
    \hline
   
   \texttt{operator} & \texttt{Infrastructure} & \texttt{Organisation} & 0.177   &0.148  &-0.082 & 0.336 & 0.332 & \textbf{0.448} \\
    \hline
    \texttt{officialLanguage} & \texttt{Populated P.} & \texttt{Language} & 0.121 &-0.041  & -0.067 & 0.691 &0.606 & \textbf{0.721}  \\
    \bottomrule
  \end{tabular}
  \caption{The transition distance scores for the properties from the DBpedia ontology.}
   \label{tab:relation}
  
\end{table*}


\paragraph{Experimental Study}
For this task, we selected 12 relations (i.e., object properties) from the DBpedia ontology along with their corresponding domain and range concepts.
Then, we measured the transition distances which are reported in 
Table \ref{tab:relation}.
The comparative results show that the GloVe model trained on Wikipedia outperforms the others. 
Interestingly, the transition distance is very high for the properties which have the shared concepts 
in the domain and range positions.



\section{Discussion and Conclusion}
\label{sec:discussion}

As it has been observed through various evaluation tasks, there is no single embedding model which shows superior performance in every scenario.
For example, while the skip-gram model performs better in the categorization task, the GloVe and CBOW model perform better for the hierarchical  task.
Thus, one conclusion is that each of these models is suited for a specific scenario.
Then, depending on the extrinsic task which consumes these embeddings, the most appropriate model should be selected.
The other conclusion is that it seems that each embedding model captures specific features of the ontological concepts,
so integrating or aligning these embeddings can be a solution for fully capturing all of these features.
Although our initial expectation was that the embeddings learned from the knowledge graph (i.e. DBpedia) should have higher quality in comparison to the embeddings learned from unstructured data (i.e. Wikipedia), in practice we did not observe that as a constant behaviour. 
We attribute this issue to two matters: (i) the weaknesses of the RDF2Vec or RDF(GloVe) approaches for generating embeddings of a knowledge graph, and (ii) the fact that Wikipedia is larger than DBpedia.
These two approaches provides a serialization on the structure of the graph (i.e. the local neighborhood of a given node is serialized) and then it runs word2vec to generate embeddings.
Here, in fact there is no discrimination between the concepts, properties, and instances, whereas the ontological resources (i.e. concepts and properties) may be required to be reinforced in the embedding model, or their embeddings have to be learned separately from the instance level.
Additionally, Wikipedia is larger than DBpedia, therefore it naturally provides richer context for the embedding models, i.e. the richer context, the higher the quality of embeddings.
Generally, we concluded that the current quality of the embeddings for ontological concepts is not in a satisfactory state.
The evaluation results are not surprising, thus providing high quality embeddings for ontological resources is an open area for future work.
Since ontological concepts play a crucial role in knowledge graphs, providing high quality embeddings for them is highly important.
We encourage the research community to utilize these metrics in their future evaluation scenarios on embedding models.
This will reduce misjudgment and provide greater insight in quality comparisons of embeddings of ontological concepts.


\bibliographystyle{aaai}
\bibliography{bib/bibEmbedding,bib/references}

\end{document}